\documentclass[journal]{IEEEtran}
\usepackage{graphicx}
\usepackage{amsmath,amssymb} 
\usepackage{color}
\usepackage{hyperref}

\begin{document}
\pagestyle{headings}

\title{A Simple Framework to Leverage State-Of-The-Art Single-Image Super-Resolution Methods to Restore Light Fields}
\author{Reuben A. Farrugia,~\IEEEmembership{Senior Member,~IEEE,}
        and~Christine Guillemot,~\IEEEmembership{Fellow,~IEEE,}
\thanks{R.A. Farrugia is with the Department
of Communications and Computer Engineering, University of Malta, Msida MSD2080, Malta, e-mail: (reuben.farrugia@um.edu.mt).}
\thanks{C. Guillemot is with is with the Institut National de Recherche en Informatique et en Automatique, Rennes 35042, France (e-mail:christine.guillemot@intria.fr).}
\thanks{Manuscript submitted July 2018. This work was supported in part by the EU H2020 Research and Innovation Programme under Grant 694122 (ERC advanced grant CLIM)}}

\maketitle
\begin{abstract}
Plenoptic cameras offer a cost effective solution to capture light fields by multiplexing multiple views on a single image sensor. However, the high angular resolution is achieved at the expense of reducing the spatial resolution of each view
by orders of magnitude compared to the raw sensor image. While light field super-resolution is still at an early stage, the field of single image super-resolution (SISR) has recently known significant advances with the use of deep learning techniques. This paper describes a simple framework allowing us to leverage state-of-the-art SISR techniques into light fields, while taking into account specific light field geometrical constraints. 
The idea is to first compute a representation compacting most of the light field energy into as few components as possible.
This is achieved by aligning the light field using optical flows and then by decomposing the aligned light field using singular value decomposition (SVD).
The \textit{principal basis} captures the information that is coherent across all the views, while the other basis contain the high angular frequencies. Super-resolving this \textit{principal basis} using an SISR method allows us to super-resolve all the information that is coherent across the entire light field. In this paper, to demonstrate the interest of the approach, we have used the very deep super resolution (VDSR) method, which is one of the leading SISR algorithms, to restore the \textit{principal basis}.
The information restored in the \textit{principal basis} is then propagated to restore all the other views using the computed optical flows.
This framework allows the proposed light field super-resolution method to inherit the benefits of the SISR method used.
Experimental results show that the proposed method is competitive, and most of the time superior, to recent light field super-resolution methods in terms of both PSNR and SSIM quality metrics, with a lower complexity.
Moreover, the subjective results demonstrate that our method manages to restore sharper light fields which enables to generate refocused images of higher quality.
\end{abstract}

\section{Introduction}

Light field imaging has recently emerged as a promising technology able to discriminate and capture light rays along different directions \cite{Ng2005,Wilburn2005}.
This rich visual description of the scene enables the creation of immersive experience in AR/VR applications and facilitates the integration of computer-generated graphics for post-production editing.
Together with proper computational algorithms, this technology is expected to impact the field of digital photography, by enabling post-capture re-focusing, depth of field extension, or 3D scene models estimation.

However, light field imaging systems trade-off spatial
resolution with angular information in the light field. Rigs of cameras capture views with a high spatial resolution but in general with limited angular sampling to reduce costs \cite{Wilburn2005}. On the other hand, plenoptic cameras
use an array of microlenses placed in front of the sensor to capture multiple low-resolution (LR) views in one 2D sensor image \cite{Ng2005}.
This is a way to cost-effectively capture multiple views with a high angular sampling, but at the expense of reducing the
spatial resolution by orders of magnitude compared to the raw sensor image.

To tackle this problem, various methods have been developed,
which are aimed to achieve better spatial and angular resolution trade-off from a plenoptic camera. These methods go from the use of coded aperture techniques, using e.g. a programmable non-refractive mask
placed at the aperture as in \cite{Liang2008}, or optically
coded projections as in \cite{Xu2012}, to light field super-resolution methods \cite{Levin2008,Bishop2012,Mitra2012,Wanner2014,Farrugia2017,Yoon2015,Yoon2017,Rossi2017}. 
While research in light field super-resolution is at its infancy, research in the related field of single-image super-resolution (SISR) is quite mature with methods based on very deep convolutional neural networks achieving state-of-the-art performances \cite{Kim2016,Kim2016b,Mao2016,Lim2017,Bae2017}.

This paper presents a framework which allows to leverage state-of-the-art 2D image super-resolution techniques to light field super-resolution. 
The energy of the light field is first compacted, to capture the coherent information,  and to then apply 2D single-image super-resolution to restore the whole light field.
To do so, we first align each view to the centre view using optical flows. 
The alignment plays an important role since it removes the disparities across the views which can be easily recovered by inverting the alignment process. 
This aligned light field is then decomposed using singular value decomposition (SVD) where the eigenvectors of the SVD capture dominant variations (or eigenimages) of the different views.
It will be shown in Section \ref{sec:lf_decomposition} that aligning the light field allows to put more information within the \textit{principal basis} (\textit{a.k.a}. dominant eigenvector) which captures the coherent information within the light field.
We then apply a state-of-the-art SISR algorithm to restore the \textit{principal basis}.
The information restored in the \textit{principal basis} is then propagated to all the other views in a consistent manner.
Inverse warping is then applied to restore the original disparities in the light field.
The results in Section \ref{sec:results} show that the proposed method achieves sharper light field images with results superior than existing methods for applications such as digital refocusing.
Supplementary material attached to this paper also show that the restored light fields are angularly coherent and that it is able to restore real-world plenoptic light fields. It is also shown that the method manages to restore light fields containing non-Lambertian surfaces\footnote{While a .ppsx file is included as supplementary material and uploaded on ScholarOne, the reviewers can watch the video at \url{https://youtu.be/HHmUZSP7HU4}}.

The main contributions of this paper are as follows:
\begin{itemize}
   \item We present a framework that enables to leverage SISR methods to restore the \textit{principal basis} capturing the coherent information across the entire light field. 
   \item Based on this framework, we describe a light field super-resolution method that yield sharper light field images with results superior than existing methods for applications such as digital refocusing.
   \item The proposed framework allows to inherit the benefits of the SISR  methods employed to restore the \textit{principal basis} and we are therefore presenting the first light field super-resolution algorithm which uses only one model to cater for different magnification factors.
\end{itemize}

The remainder of this paper is organized as follows. Work related to the method described in this paper is provided in Section  \ref{sec:related_work} while the light field energy compaction method is explained in Section \ref{sec:lf_decomposition}.
The proposed \textit{principal basis} VDSR (PB-VDSR) is described in Seciton \ref{sec:principal_basis_SR} while the experimental results are delived in the following section.
Section \ref{sec:conclusion} concludes with the final remarks.

\section{Related Work}
\label{sec:related_work}

This section gives a brief overview of work related to the key concepts of the proposed spatial light field super-resolution approach and the light field super-resolution methods that are found in literature.

\subsection{Single Image Super-Resolution}
\label{sec:sisr_related_work}

Single-image super-resolution is an ill-posed inverse problem with infinite possible solutions.
These methods use priors to derive a more plausible solution that satisfies a predefined assumption. 
These priors are either hand-crafted, such as total variation or Bayesian models, or data driven that are learned using machine learning methods.
Pixel-based methods have been proposed in \cite{HeSiu2011}, \cite{ZhangGao2012} where each pixel in the high-resolution (HR) image is inferred via statistical learning. To improve spatial coherency, patch-based approaches, referred to as example-based methods, have been proposed.
Freeman \emph{et. al}. \cite{Freeman2002} presented the first single-image example-based super-resolution algorithm that used a coupled dictionary to learn a mapping between LR and HR patches.
More advanced methods based on manifold learning \cite{Chang2004,Gao2012,ferreira:hal-01388955} and sparse coding \cite{Yang2012,Timofte2015} were investigated to regularize the problem and were found to provide sharper images. 
Other approaches \cite{Glasner2009,Freedman2011,Yang2013,bevilacqua:hal-01088753} utilized image self-similarities to avoid using dictionaries constructed using external images.

Deep neural networks have contributed to a drastic improvement in the field of single-image super-resolution.
Dong \emph{et. al}. \cite{Dong2014} were the first to use a rather shallow convolutional neural network (SRCNN).
Residual learning was introduced in \cite{Kim2016,Kim2016b,Bae2017} for training deeper network architectures and achieved state-of-the-art performance. The authors in \cite{Mao2016} pose the general image restoration problem with encoder-decoder networks and systematic skip connections.
This architecture was later on extended in \cite{Lim2017} where the authors expanded the model size and removed unnecessary modules in the convolutional residual networks.

\subsection{Light Field Super-Resolution}
\label{sec:lf_SR_related_work}

Early light field super-resolution approaches pose the  problem as one of recovering the high-resolution views from multiple low-resolution images with unknown non-integer translation misalignment.
The authors in \cite{Levin2008,Bishop2012} proposed a two-step approach where they first estimate a depth map and then formulate the super-resolution problem either as a simple linear problem \cite{Levin2008} or as a Beyesian inference problem \cite{Bishop2012} assuming an image formation model with Lambertian reflectance priors and depth-dependent blurring kernels.
A patch-based technique was proposed in \cite{Mitra2012} where  high-resolution 4D patches are estimated using a linear minimum mean square error (LMMSE) estimator assuming a disparity-dependent Gaussian Mixture Model (GMM) for the patch structure.
A variational optimization framework was proposed in \cite{Wanner2014} to spatially super-resolve the light field given their estimated depth maps and to increase the angular resolution.

Example-based light field super-resolution methods have been recently proposed. These methods use machine learning to learn a mapping between low- and high-resolution light fields.
In \cite{Farrugia2017}, the authors show that a 3D patch-volume resides on a low-dimensional subspace and propose to learn a projection between low- and high-resolution subspaces of patch-volumes using ridge-regression.
Deep learning techniques for light field super-resolution have been first proposed in \cite{Yoon2015} where 4-tuples of neighbouring views are stacked into groups and restored using SRCNN \cite{Dong2014}. 
The spatially restored light field is then fed into a second CNN that up-scales the angular resolution.
The same authors have later proposed to restore each view independently using SRCNN in \cite{Yoon2017} showing superior performance over their original method.
More recently, graph based light field super-resolution algorithm was presented in \cite{Rossi2017} that enforces the optimization to preserve the light field structure. 
A shallow neural network was proposed in \cite{Gul2018} to restore light fields captured by a plenoptic camera.
However, this method is only suitable to achieve a magnification factor of $\times 2$ and needs to train a CNN for every angular view.
Very recently, a multi-scale fusion scheme was used to accumulate contextual information from multiple scales while Recurrent Convolutional Neural Networks (BRCNN) is used to model the spatial relation between adjacent views and restore the light field.

A hybrid light field super-resolution method was proposed in \cite{Wang2017} where a high-resolution camera was coupled with a plenoptic camera.
The authors in \cite{Wang2017b} describe an acquisition device formed by eight low-resolution side cameras arranged around a central high-quality camera.
Iterative patch- and depth-based synthesis (iPADS) is then used to reconstruct a light field with the spatial resolution of the SLR camera and an increased number of views.

While the methods in \cite{Yoon2015, Yoon2017,Gul2018,Wang2018} use deep learning to super-resolve the light field, our method is considerably different.
The novelty of our approach is that the proposed framework allows to use SISR techniques for light field super-resolution.
The deep learning SR method used are not retrained on light fields and use models that are trained on natural images.
Moreover, our framework inherits the benefits of the SISR algorithm used.
In our study, we used VDSR \cite{Kim2016} which allows us to use a very deep super-resolution method that adopts one single model to cater for different magnification factors.

\subsection{Light Field Edit Propagation}
\label{sec:lf_edit_propagation}

Light field edit propagation involves the restoration of the centre view followed by the propagation of the restored information to all the other views.
The authors in \cite{Seitz1998} described an approach using a 3D voxel-based model of the scene with an associated radiance function to propagate pixel edits and illumination changes in a consistent manner from one view to the other views of the light field.
The authors in \cite{Jarabo2011} extend the 2D image stroke-based edit propagation method of \cite{An2008} to light fields, where they reduce the complexity by propagating the edits in a downscaled version of the light field.
In \cite{Ao2015}, a method based on a reparameterization of the light field is proposed to better preserve coherence of the edits in the angular domain.
However, these methods deal with simple stroke-based editing and are not suitable to propagate complex edits, such as inpainting or super-resolution, to all the other views.

A patch-based depth-layer-aware image synthesis algorithm was adopted in \cite{Zhang2017} to propagate the edits from the centre view to all the other views.
The authors in \cite{Frigo2017} use tensor driven diffusion to propagate information from the centre view along the Epipolar Plane Image (EPI) structure of the light field.
These methods were used to propagate either simple edits, recolorization or inpainting from the center view to all the other views.
However, up to the knowledge of the authors, such approaches were never been considered for light field super-resolution.

\section{Light Field Energy Compaction}
\label{sec:lf_decomposition}

Lets consider an input light field $I(x,y,s,t)$ represented with the two plane parametrization proposed in \cite{Levoy1996,Gortler1996}, where $(x,y)$ and $(s,t)$  represent spatial and angular coordinates respectively.
The light field can be seen as a 2D array of images, where each image $\mathbf{I}_{s,t}$ captures the scene from a viewpoint defined by angular coordinates $(s,t)$.
One can use single image super-resolution technique to restore every angular view independently.
However, these methods do not exploit the geometrical structure of the light field \cite{Liang2015} and are not guaranteed to provide angularly coherent solutions \cite{Farrugia2017}.
On the other hand, several light field super-resolution techniques have been proposed that either exploit the disparity/depth information \cite{Levin2008,Bishop2012,Mitra2012,Wanner2014}  or else use learning based methods \cite{Farrugia2017,Yoon2015,Yoon2017,Gul2018,Wang2018} to improve the quality of the light field.
However, these algorithms do not benefit from the recent advances in single image super-resolution where very deep Convolutional Neural Networks are achieving outstanding performances \cite{Dong2014,Kim2016,Kim2016b,Lim2017,Mao2016}.

\begin{figure*}[htb]
\begin{minipage}[b]{0.5\linewidth}
	\centering 
	\begin{tabular}{ccc}
		\includegraphics[width=0.3\linewidth]{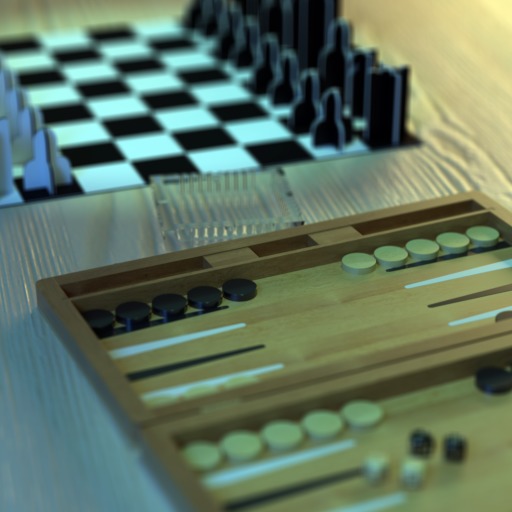} & 					\includegraphics[width=0.3\linewidth]{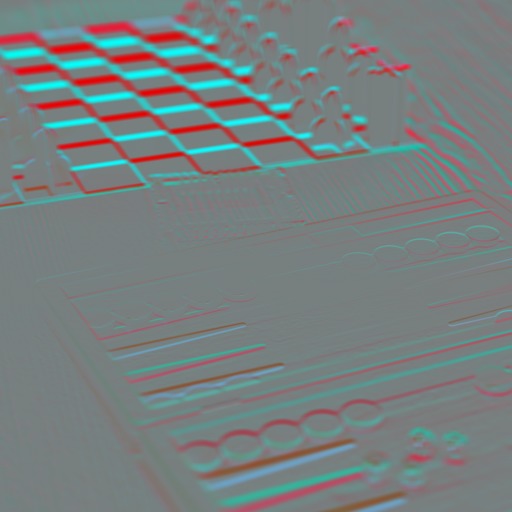} & \includegraphics[width=0.3\linewidth]{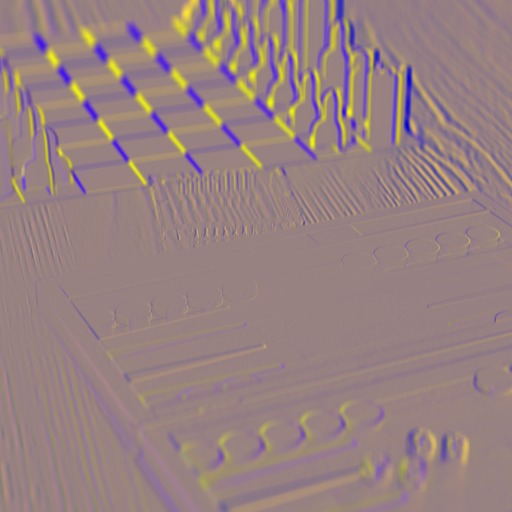}  \\
	$\mathbf{B}_0: 7.66$ & $\mathbf{B}_1: 5.45$ & $\mathbf{B}_2: 5.50$  \\
	\includegraphics[width=0.3\linewidth]{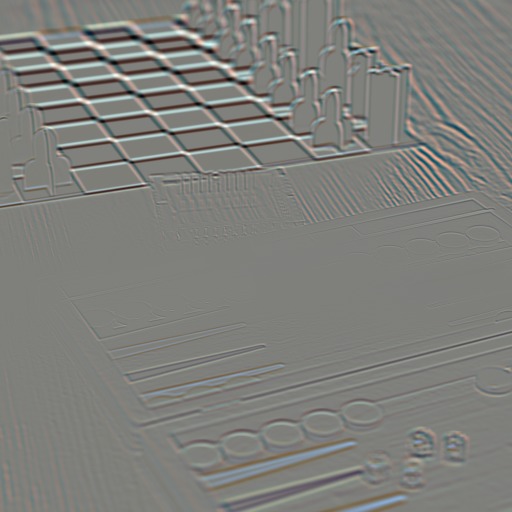} & \includegraphics[width=0.3\linewidth]{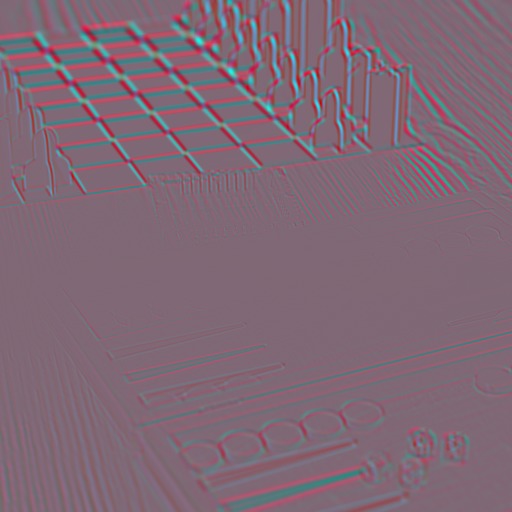} & \includegraphics[width=0.3\linewidth]{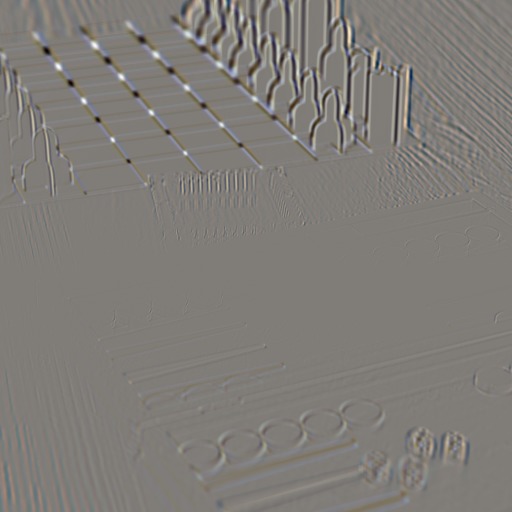}  \\
		$\mathbf{B}_3: 5.09$ & $\mathbf{B}_4: 5.64$ & $\mathbf{B}_5: 4.72$  \\
	\end{tabular}  
\end{minipage} 
\begin{minipage}[b]{0.5\linewidth}
	\centering 
	\begin{tabular}{ccc}
		\includegraphics[width=0.3\linewidth]{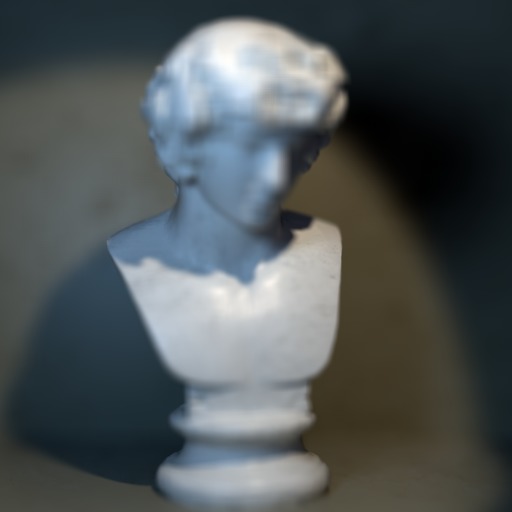} & 					\includegraphics[width=0.3\linewidth]{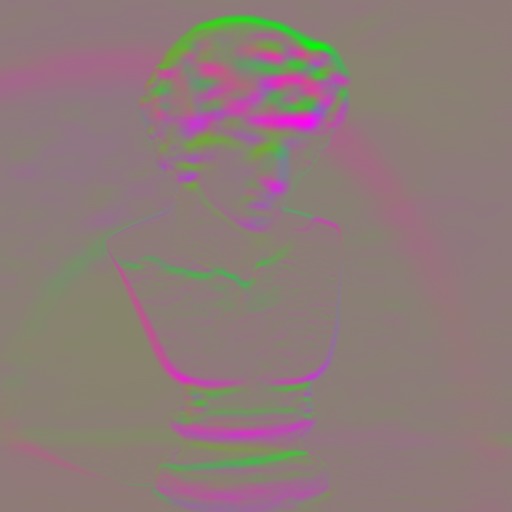} & \includegraphics[width=0.3\linewidth]{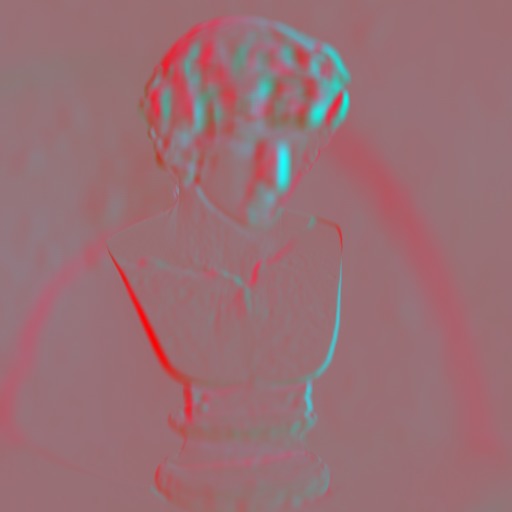}  \\
	$\mathbf{B}_0: 7.56$ & $\mathbf{B}_1: 5.76$ & $\mathbf{B}_2: 5.71$  \\
	\includegraphics[width=0.3\linewidth]{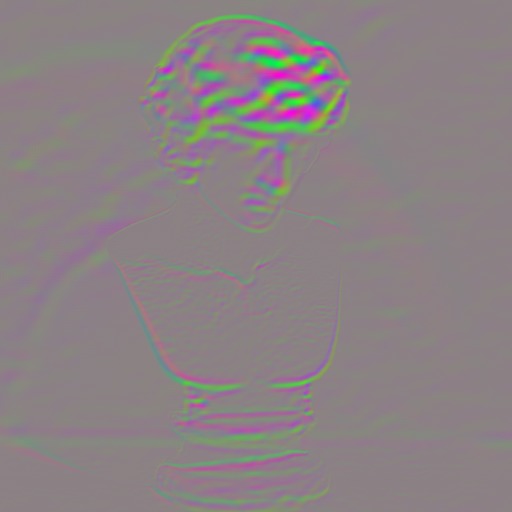} & \includegraphics[width=0.3\linewidth]{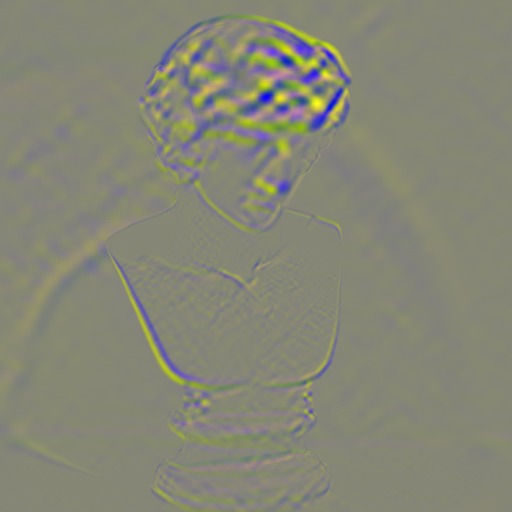} & \includegraphics[width=0.3\linewidth]{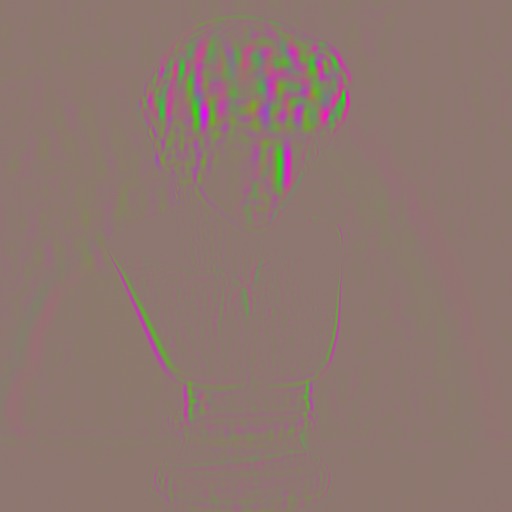}  \\
		$\mathbf{B}_3: 4.95$ & $\mathbf{B}_4: 4.90$ & $\mathbf{B}_5: 5.06$  \\
	\end{tabular}  
\end{minipage} 
\centerline{(a) First six-basis computed using SVD.}\\

\begin{minipage}[b]{0.48\linewidth}
	\centering 
	\begin{tabular}{ccc}
		\includegraphics[width=0.3\linewidth]{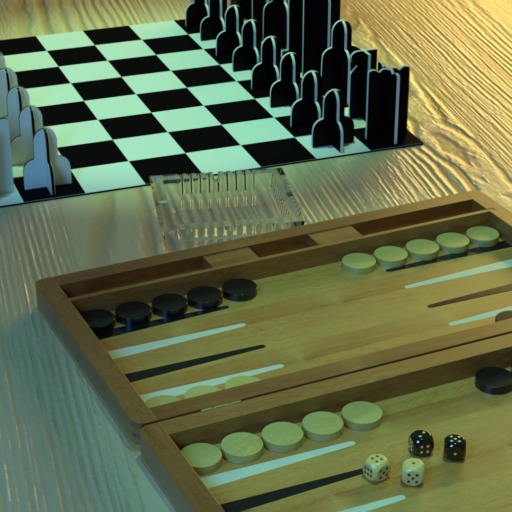} & 					\includegraphics[width=0.3\linewidth]{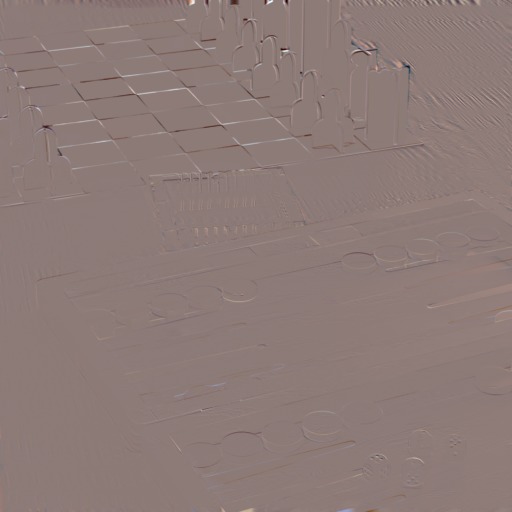} & \includegraphics[width=0.3\linewidth]{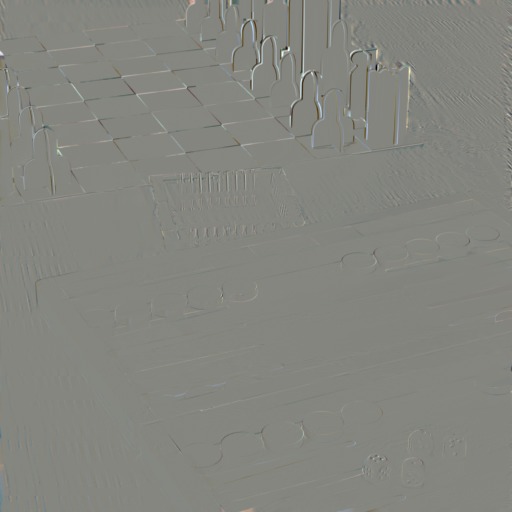}  \\
	$\mathbf{B}_0: 7.55$ & $\mathbf{B}_1: 4.17$ & $\mathbf{B}_2: 3.75$  \\
	\includegraphics[width=0.3\linewidth]{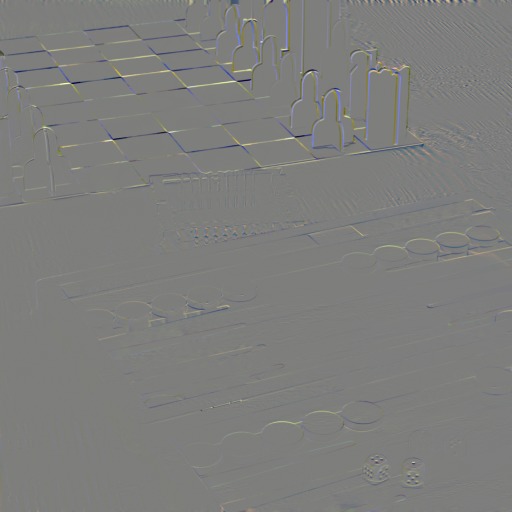} & \includegraphics[width=0.3\linewidth]{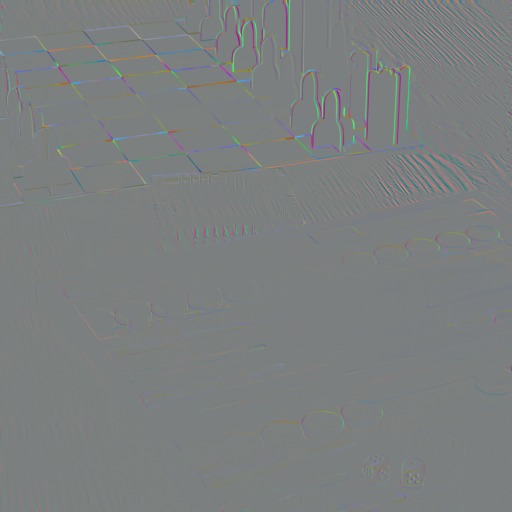} & \includegraphics[width=0.3\linewidth]{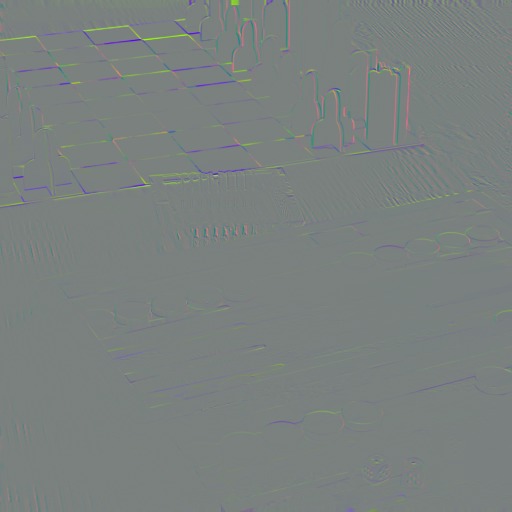}  \\
		$\mathbf{B}_3: 3.21$ & $\mathbf{B}_4: 3.48$ & $\mathbf{B}_5: 3.67$  \\
	\end{tabular}  
\end{minipage} 
\begin{minipage}[b]{0.48\linewidth}
	\centering 
	\begin{tabular}{ccc}
		\includegraphics[width=0.3\linewidth]{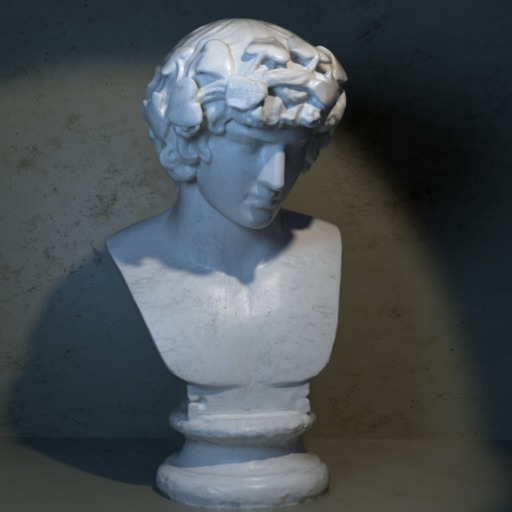} & 					\includegraphics[width=0.3\linewidth]{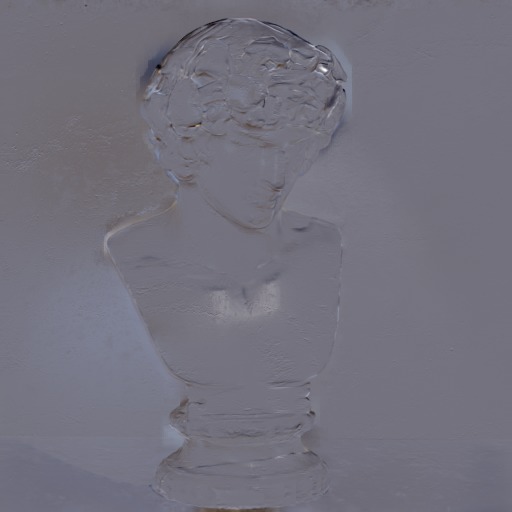} & \includegraphics[width=0.3\linewidth]{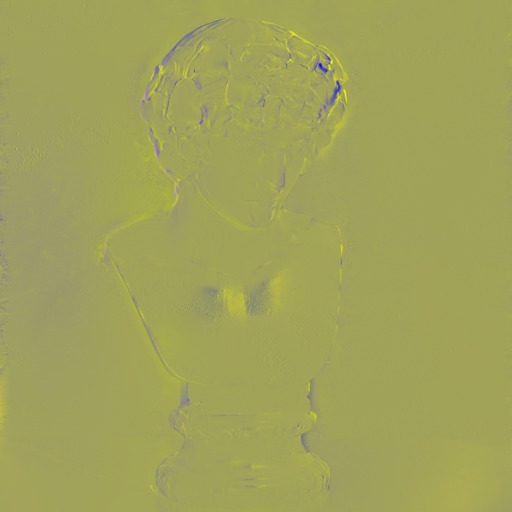}  \\
	$\mathbf{B}_0: 7.29$ & $\mathbf{B}_1: 4.88$ & $\mathbf{B}_2: 4.71$  \\
	\includegraphics[width=0.3\linewidth]{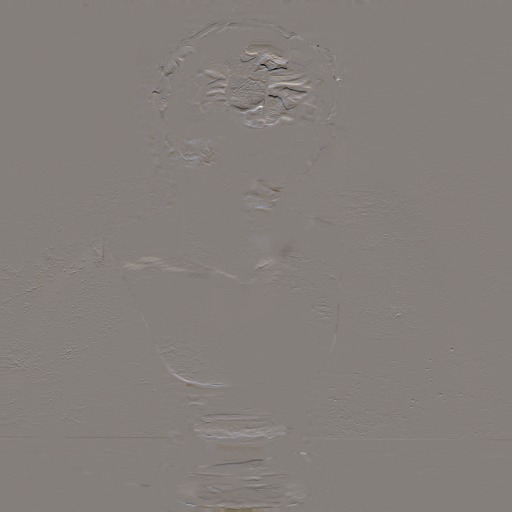} & \includegraphics[width=0.3\linewidth]{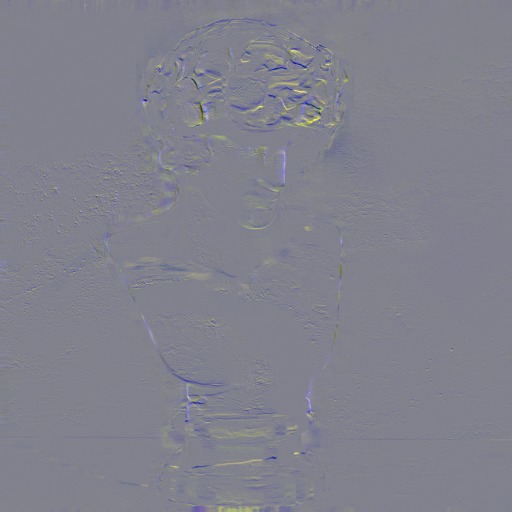} & \includegraphics[width=0.3\linewidth]{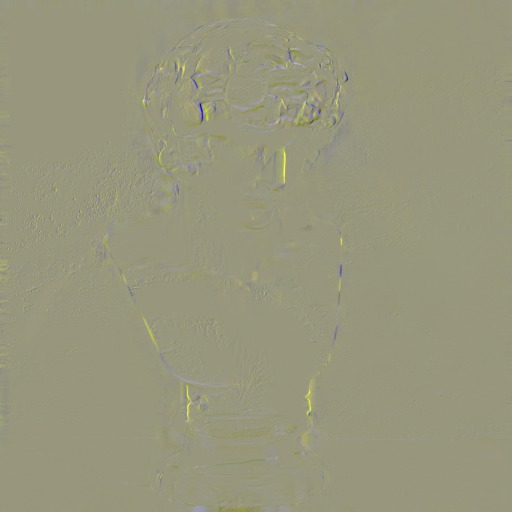}  \\
		$\mathbf{B}_3: 3.94$ & $\mathbf{B}_4: 4.30$ & $\mathbf{B}_5: 4.11$  \\
	\end{tabular}  
\end{minipage} 
\centerline{(b) First six-basis computed using SVD of the aligned light field.}\\
\caption{Comparing the information contained in each basis when computing (a) SVD decomposition and (b) Aligned SVD decomposition for the (left) Boardgames and (right) Antinous synthetic light fields.
The entropy measure for each basis is given below the corresponding image.}
\label{fig:svd_decomposition2}
\end{figure*}

A light field consists of a very large volume of high-dimensional data.
Nevertheless, it exhibits redundancies in all four dimensions since every view captures the same scene from a slightly different viewpoint.
Early work in the field of light field compression used 3D/4D wavelet transforms to decompose the light field into a number of sub--bands \cite{Peter99,Lalonde1999,Chang2006}, where each sub--band gives information at different spatial and angular frequencies.
Figure \ref{fig:svd_decomposition2}(a) shows the the first six orthogonal basis when decomposing the light field using SVD.
It can be seen that while most of the energy resides in the \textit{principal basis} $\mathbf{B}_0$, there is still a lot of high frequency detail in the other basis.
Moreover, the \textit{principal basis} $\mathbf{B}_0$, which captures the average energy in the scene is blurred.
This is attributed to variations in disparities across the views which result in high-frequency angular details that are not captured by the \textit{principal basis}.

\begin{figure*}[ht]
\centering
\begin{tabular}{cccccc}
\centering

\footnotesize{{No Align}} & \footnotesize{{RASL}} \cite{Peng2010} &  \footnotesize{HLRA \cite{Jiang2017}} & \footnotesize{SIFT Flow \cite{Liu2011}} & \footnotesize{CPM \cite{Hu2016}} & \footnotesize{SPM-BP \cite{Li2015}} \\ 
\includegraphics[width=2.5cm]{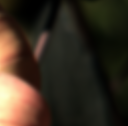} & 
\includegraphics[width=2.5cm]{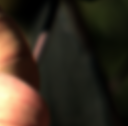} & 
\includegraphics[width=2.5cm]{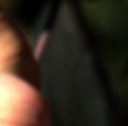}
&
\includegraphics[width=2.5cm]{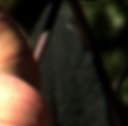} 
&
\includegraphics[width=2.5cm]{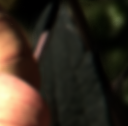} 
&
\includegraphics[width=2.5cm]{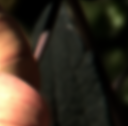} 
\\
\footnotesize{99.092} & \footnotesize{98.261} & \footnotesize{99.080} &  \footnotesize{19.164} & \footnotesize{51.073} & \footnotesize{71.652} \\ 
\includegraphics[width=2.5cm]{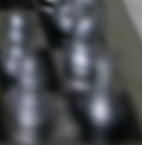} & 
\includegraphics[width=2.5cm]{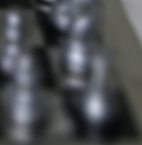} & 
\includegraphics[width=2.5cm]{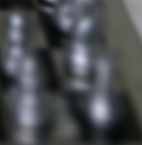}
&
\includegraphics[width=2.5cm]{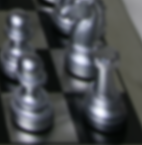} 
&
\includegraphics[width=2.5cm]{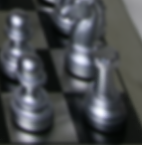} 
&
\includegraphics[width=2.5cm]{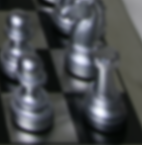} 
\\
\footnotesize{102.625} & \footnotesize{100.529} & \footnotesize{102.478} &  \footnotesize{3.595} & \footnotesize{23.978} & \footnotesize{4.638} \\ 
\includegraphics[width=2.5cm]{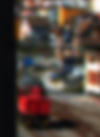} & 
\includegraphics[width=2.5cm]{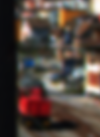} & 
\includegraphics[width=2.5cm]{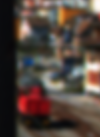}
&
\includegraphics[width=2.5cm]{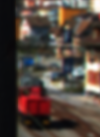} 
&
\includegraphics[width=2.5cm]{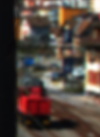} 
&
\includegraphics[width=2.5cm]{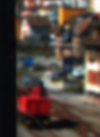} 
\\
\footnotesize{76.039} & \footnotesize{75.150} & \footnotesize{75.944} &  \footnotesize{12.973} & \footnotesize{18.815} & \footnotesize{15.100} \\ 
\end{tabular} 
\caption{Cropped regions of the mean view when using different disparity compensation methods. Underneath each image we provide the average variance across the $n$ angular views which was used in \cite{Farrugia2017} to characterize the performance of the alignment algorithm, where smaller values indicate  better alignment.  }
\label{fig:subjective_Eval1}
\end{figure*}

The authors in \cite{Jiang2017} tried to reduce the energy within the high-frequency basis by jointly aligning the angular views and estimating a low--rank approximation (LRA) of the light field.
This approach has shown very promising results in the field of light field compression.
In the same spirit, the RASL algorithm \cite{Peng2010} was used to find the homographies that globally align a batch of linearly correlated images.
Both methods find an optimal set of homographies such that the matrix of aligned images can be decomposed in a low--rank matrix of aligned images, with the latter constraining the error matrix to be sparse.
However, as it can be seen in Figure \ref{fig:subjective_Eval1}, while both RASL and HLRA methods manage to globally align the angular views, the resulting mean view, that is computed by averaging all the views, are still blurred indicating that the views are not well aligned. 

In the sequel, we consider $\mathbf{I}_{s,t}$ to represent different views, where $(s,t)$ define the angular coordinates. 
This notation will be further simplified as $\mathbf{I}_i$ with a bijection between $(s,t)$ and $i$. 
The complete light field can hence be represented by a matrix $\mathbf{I} \in \mathbb{R}^{m,n} $:

\begin{equation}
\mathbf{I} = [vec(\mathbf{I}_{1}) \quad | \quad vec(\mathbf{I}_{2}) \quad | \quad \cdots \quad | \quad vec(\mathbf{I}_{n})]
\label{eq:lf_matrix}
\end{equation}

\noindent with $vec(\mathbf{I}_i)$ being the vectorized representation of the $i$-th angular view, $m$ represents the number of pixels in each view $(m = X \times Y)$ and $n$ is the number of views in the light field $(n = P \times Q)$, where $P$ and $Q$ represent the number of vertical and horizontal angular views respectively.
We then formulate the light field decomposition problem as that of finding a set of orthogonal basis $\mathbf{B}$ that is able to capture most of the information contained in the light field.
This can be achieved by minimizing the following optimization problem

\begin{equation}
\label{eq:lf_decomposition}
\underset{\mathbf{u},\mathbf{v},\mathbf{B},\mathbf{C}}{
min}_{}{ || \Gamma_{\mathbf{u},\mathbf{v}} \left( \mathbf{I}\right) - \mathbf{B}\mathbf{C}||^2_2 } 
\end{equation}

\noindent where $\mathbf{u} \in \mathbb{R}^{m,n}$ and $\mathbf{v} \in \mathbb{R}^{m,n}$ are flow vectors that specify the displacement of each pixel needed to align each view with the centre view, $\mathbf{B} \in \mathbb{R}^{m,n}$ represents the basis matrix, $\mathbf{C} \in \mathbb{R}^{n,n}$ is the combination weight matrix and $\Gamma_{\mathbf{u},\mathbf{v}}(\cdot)$ is the forward warping operator.

This optimization problem is computationally intractable.
Instead, we decompose this problem in two sub--problems: \romannumeral 1) use an optical flow estimation technique to find the flow vectors $\mathbf{u}$ and $\mathbf{v}$ that best align each view with the centre view and \romannumeral 2) decompose the aligned light field into a set of basis $\mathbf{B}$ and coefficient matrix $\mathbf{C}$ using SVD.
The results in Figure \ref{fig:subjective_Eval1} clearly show that the mean views are much sharper when aligning the light field using optical flows. 
Moreover, optical flows significantly reduce the variance across the angular views, with the SIFT flow method \cite{Liu2011} achieving the best performance. It reduces the mean variance across views by a factor of nine, and thus we will use it to align the views.
Reducing the total variance across the views (as shown in Figure \ref{fig:subjective_Eval1}) allows to compact more information in the low-frequency basis.

The solution of the first sub-problem gives the flow-vectors $\mathbf{u}$ and $\mathbf{v}$ which are used to align the light field using forward warping \emph{i.e}. $\tilde{\mathbf{I}} = \Gamma_{\mathbf{u},\mathbf{v}} \left( \mathbf{I}\right)$.
The aligned light field $\tilde{\mathbf{I}} = \mathbf{U} \boldsymbol{\Sigma} \mathbf{V}^T$ is then decomposed using SVD, where $\mathbf{U}$ and $\mathbf{V}$ are unitary matrices and $\boldsymbol{\Sigma}$ is a diagonal matrix containing the singular values.
The basis matrix is computed as $\mathbf{B} = \mathbf{U} \boldsymbol{\Sigma}$ while the coefficient matrix is given by $\mathbf{C}= \mathbf{V}^T$.

Figure \ref{fig:svd_decomposition2}(b) shows the first six orthogonal basis using our proposed light field decomposition method.
It can be seen that the \textit{principal basis} $\mathbf{B}_0$ is much sharper indicating that it captures more information from the light field.
Moreover, the energy in the higher-frequency basis is significantly reduced as indicated by the significant drop in entropy when using our proposed light field decomposition method.
The light field can then be easily reconstructed using $\tilde{\mathbf{I}} = \mathbf{B} \mathbf{C}$ without losing any information since $\mathbf{B}$ is orthogonal and full-rank. 
In the sequel, the decompositon of the aligned light field will be referred to as A-SVD.

\section{Principal Basis Super-Resolution}
\label{sec:principal_basis_SR}

\begin{figure*}[ht]
\centering
\def\svgwidth{\linewidth} 
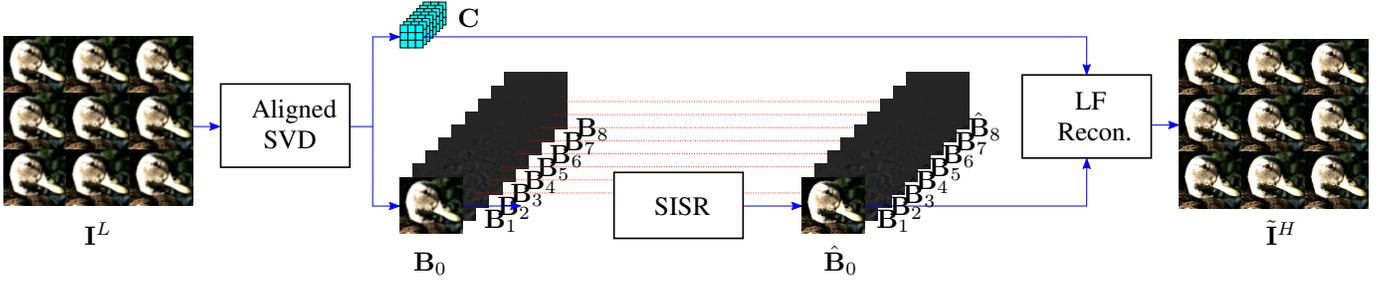
\caption{The proposed light field super-resolution algorithm that takes a $3 \times 3$ matrix of low-resolution views as input, denoted by $\mathbf{I}^L$, to estimate the high-resolution light field $\mathbf{I}^H$.}
\label{fig:pb_vdsr_block_diagram}
\end{figure*}

Let $\mathbf{I}^H$ and $\mathbf{I}^L$ denote the high- and low-resolution light fields. The super-resolution problem can be formulated in Banach space as

\begin{equation}
\mathbf{I}^L = \downarrow_\alpha \mathbf{I}^H + \boldsymbol{\eta}
\label{eq:acquision_model}
\end{equation}

\noindent where $\boldsymbol{\eta}$ is an additive noise matrix and $\downarrow_\alpha$ is a downsampling operator applied on each angular view with a scale--factor $\alpha$.
Figure \ref{fig:pb_vdsr_block_diagram} illustrates a block diagram of the proposed light field super-resolution algorithm where for simplicity a $3 \times 3$ matrix of angular views is shown.
The A-SVD algorithm, described in section \ref{sec:lf_decomposition}, is applied on the low-resolution light field $\mathbf{I}^L$ to decompose the light field into a set of orthogonal basis $\mathbf{B} \in \mathbb{R}^{m,n}$ and coefficient matrix $\mathbf{C} \in \mathbb{R}^{n,n}$.
As shown in more detail in Section \ref{sec:lf_decomposition}, the A-SVD algorithm is able to capture more information in the \textit{principal basis} $\mathbf{B}_0$.

Driven by the observation that the \textit{principal basis} $\mathbf{B}_0$ is a natural image that captures most of the information in the light field, we pose the problem of light field super-resolution as that of restoring the resolution of the \textit{principal basis}.
The higher order basis $\mathbf{B}_j$, $j \in [1,n-1]$, that capture the discrepancies across the views in terms of occlusions and illumination, are not modified.
Any single SISR method can be used to restore the low resolution \textit{principal basis} $\mathbf{B}_0$ and to estimate the high resolution \textit{principal basis} $\hat{\mathbf{B}}_0$.
The high frequency basis are simply approximated using $\hat{\mathbf{B}}_j = \mathbf{B}_j \text{ for } j \in [1,n-1]$.
The restored aligned light field is then reconstructed using a simple matrix multiplication which is then inverse-warped to restore the original disparities \textit{i.e}. $\tilde{\mathbf{I}}^H = \Gamma^{-1}_{\mathbf{u},\mathbf{v}} (\hat{\mathbf{B}} \mathbf{C})$, where $\Gamma^{-1}_{\mathbf{u},\mathbf{v}}(\cdot)$ stands for the inverse warping operator.
While the theoretical and implementation details of A-SVD were provided in Section \ref{sec:lf_decomposition}, the following sub-sections will deal with the implementation detail of the SISR and LF Reconstruction modules.

\subsection{SISR Module}
\label{sec:restore_pb}

In this work we consider some of the most promising SISR methods found in literature to restore the \textit{principal basis} and their performance is summarized in Table \ref{tbl:SISR_performance}.
In essence we consider the first deep-learning based super-resolution method SRCNN \cite{Dong2014}, the very deep convolutional neural network (VDSR) which uses residual learning with 20 convolutional layers \cite{Kim2016} and the Lab402 method which was ranked third in the recent NTIRE workshop challenge. 
The network models of these methods were not retrained on light field data and therefore this experiment evaluates the generalization abilities of these methods.
These results demonstrate that while both VDSR and Lab402 manage to outperform SRCNN, the VDSR method is able to achieve the best performance in terms of both PSNR and SSIM quality measures.
This indicates that while other methods can be used to restore the \textit{principal basis}, the VDSR algorithm achieves the best performance and will therefore be considered in the experimental results in Section \ref{sec:results}.
Given that our method uses VDSR to restore the \textit{principal basis} we named our method PB-VDSR.
It is important to mention here that unlike SRCNN, VDSR uses a single network model to cater for different magnification factors and PB-VDSR inherits this property.

\begin{table}[htb]
\caption{Quality analysis (PSNR with SSIM in parenthesis) using different single-image super-resolution algorithms to restore the \textit{principal basis} $\mathbf{B}_0$ at a magnification factor $\times 3$.}
\label{tbl:SISR_performance}
\begin{center}
\begin{tabular}{|l|c|c|c|}
\hline
\bf{Light Field} & \bf{SRCNN} \cite{Dong2014} & \bf{VDSR} \cite{Kim2016} & \bf{Lab402} \cite{Bae2017}  \\ 
\hline
Antinous     & 33.32 (0.954) & \bf{35.74} (\bf{0.978}) & 33.81 (0.977) \\
Boardgames   & 23.68 (0.835) & \bf{24.65} (\bf{0.865}) & 23.92 (0.859)\\
Greek        & 30.78 (0.935) & \bf{33.55} (\bf{0.966}) & 31.70 (0.961) \\
Medieval 2   & 30.32 (0.952) & \bf{32.10} (\bf{0.962}) & 31.74 (\bf{0.962})\\
Origami      & 25.32 (0.951) & 28.89 (\bf{0.973}) & \bf{28.97} (\bf{0.973})\\
Books       & 29.73 (0.966) & \bf{30.78} (0.974) & 29.86 (0.970))\\
Friends 2   & 29.31 (0.935) & \bf{31.13} (\bf{0.944}) & 30.79 (\bf{0.944}) \\
Game Board & 31.75 (0.972) & \bf{32.12} (\bf{0.976}) & 31.54 (0.974) \\
Graffiti   & 28.56 (0.870) & 29.90 (0.880) & \bf{29.91} (\bf{0.883}) \\
Parc du Luxembourg & 28.49 (0.926) & \bf{29.08} (\bf{0.935}) & 28.39 (0.928) \\
\hline
\end{tabular}
\end{center}
\end{table}

\subsection{Light Field Reconstruction Module}
\label{sec:lf_reconstruct}

\begin{figure}[htb]
\centering
\begin{tabular}{ccc}
\includegraphics[width=0.3\linewidth]{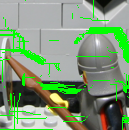} &
\includegraphics[width=0.3\linewidth]{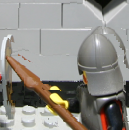} &
\includegraphics[width=0.3\linewidth]{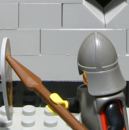} \\
\includegraphics[width=0.3\linewidth]{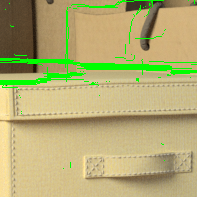} &
\includegraphics[width=0.3\linewidth]{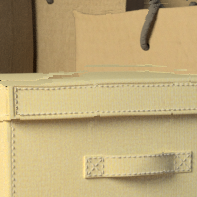} &
\includegraphics[width=0.3\linewidth]{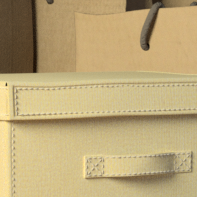} \\
\footnotesize{Forward Warping} & \footnotesize{Inverse Warping} & \footnotesize{Proposed Inpainting}\\ 
\end{tabular} 
\caption{Inpainting the cracks marked in green}
\label{fig:inpainting}
\end{figure}

The aligned high resolution light field can be estimated by multiplying the restored basis $\hat{\mathbf{B}}$ and weight matrix $\mathbf{C}$ \textit{i.e}. $\hat{\mathbf{I}}^H = \hat{\mathbf{B}} \mathbf{C}$.
The views of $\hat{\mathbf{I}}^H$ are aligned with the center view.
Forward warping can be used to recover the original disparities of the restored views.
However, as can be seen in the first column of Figure \ref{fig:inpainting}, forward warping is not able to restore all pixels and results in a number of cracks and holes.
Another approach is to use inverse warping and use neighbouring pixels to estimate the missing information.
However, as can be seen in the second column of Figure \ref{fig:inpainting}, missing pixels due to occlusion are not well correlated with the neighbouring pixels and  result in inaccurate estimates.

In this work we observe that the pixels warped using forward warping are very accurate.
Instead of interpolating the missing pixels, in this work we simply copy the collocated pixels from the low-resolution light field to replace the missing pixels.
The light field reconstructed in this work is depicted as the third column of Figure \ref{fig:inpainting} where it can be seen that the recovered pixels are more accurate than those estimated using inverse warping. 

\subsection{Edit Propagation Methods}
\label{sec:edit_prop_compare}

\begin{figure*}[ht]
\centering
\setlength\tabcolsep{1.5pt}
\begin{tabular}{cc}
\centering
\includegraphics[width=0.4\linewidth]{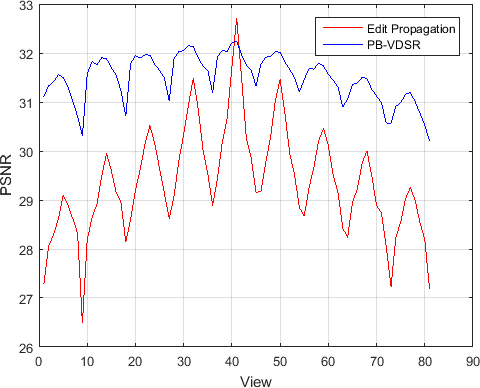} & 
\includegraphics[width=0.4\linewidth]{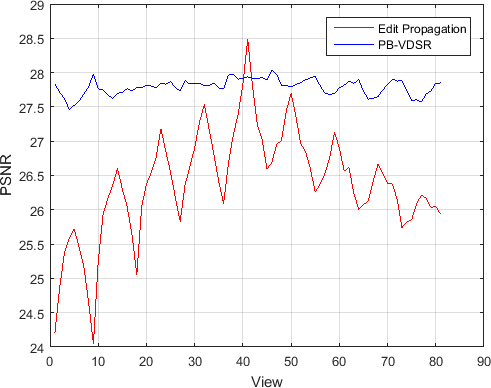}  \\
Bee 2 (INRIA) & Bikes (EPFL) \\
\end{tabular}
\caption{PSNR analysis at each view comparing the proposed PB-VDSR to an edit propagation method. The view index indicates the index of the view when scanned using raster scan ordering.}
\label{fig:edit_propagation_analysis_}
\end{figure*}

This work is related to the Light Field Edit Propagation methods described in Section \ref{sec:lf_edit_propagation} which allow the user to edit the center view and propagate the edits to all the other views.
Figure \ref{fig:edit_propagation_analysis_} illustrates the PSNR measure at each view and compares the proposed method, which we call PB-VDSR, against the edit propagation method that will be described next.
The edit propagation considered applies VDSR to restore the center view and then propagate the information to the other views using forward warping.
The missing pixels due to occlusions are estimated using  collocated pixels from the low-resolution light field as described in the previous subsection.

It can be seen that edit propagation achieves larger PSNR for the center view (view 41). However its performance degrades significantly when propagating the information to all the other views. 
This can be explained since the edit propagation ignores the variations across the views caused by illumination and occlusions.
On the other hand, PB-VDSR restores the \textit{principal basis} that captures the angular consistent information in the light field while the variations caused by illumination and occlusion are preserved in the higher frequency basis.
This implies that PB-VDSR propagates the high angular frequency information in $\hat{\mathbf{I}}^H$.
Moreover, the higher PSNR achieved at the center view by the edit propagation is obtained using the VDSR network  trained to restore natural images like the center view and not the \textit{principal basis} which captures the dominant information in the light field.
This result suggests that the performance of the algorithm can be further improved by retraining the VDSR neural network to specifically restore \textit{principal basis} rather than considering it as a generic image. 
However, retraining the SISR is not in scope of this paper since the objective here is to show that SISR can be extended using our framework to restore light fields.

\section{Experimental Results}
\label{sec:results}

The experiments conducted in this paper use real-world light fields from the EPFL \cite{Rerabek2016}, INRIA\footnote{INRIA dataset: https://www.irisa.fr/temics/demos/IllumDatasetLF/index.html},  and Stanford\footnote{Stanford dataset: http://lightfield.stanford.edu/} datasets.
Both EPFL and INRIA are light fields that are captured by a plenoptic camera and therefore have low angular disparities while the Stanford dataset is captured using a Gantry which have larger angular disparities.
While the angular views of the EPFL and Stanford datasets are available, the light fields in the INRIA dataset were decoded using the method in \cite{Dansereau2013}.
In all our experiments we consider a $9 \times 9$ matrix of angular views.
For computational purposes, the high-resolution views of the Stanford dataset were down-sampled such that the lowest dimension is set to 400 pixels.
The high-resolution images of the other datasets were kept unchanged \textit{i.e}. $625 \times 434$.

We compare the performance of our proposed PB-VDSR method against some of the best performing methods in the field of light field super-resolution, namely  the CNN based light field super-resolution algorithm (LF-SRCNN) \cite{Yoon2017}, the linear subspace projection based method (BM-PCARR) \cite{Farrugia2017} and the Graph-based light field super resolution (GRAPH) \cite{Rossi2017}.
It must be mentioned that while the BM+PCARR and LF-SRCNN were retrained on 98 light fields that were not considered in the evaluation phase, the network model adopted by VDSR was not retrained on light fields and we used the original model adopted for single image super-resolution.
Moreover, PB-VDSR inherits the benefits of VDSR and adopts one single model to cater for different magnification factors.
The other light field super-resolution methods described in the related work section were not considered since they either were reported to achieve performance inferior to the methods considered here \cite{Yoon2017,Farrugia2017,Rossi2017} or the code was not made publicly available by the authors at the time of writing the paper.
The MATLAB code of the proposed method will be made available online upon publication\footnote{LF-Editing Repository: https://github.com/rrfarr/LF-Editing}.

\begin{table*}[ht]
\caption{Quality analysis (PSNR with SSIM in parenthesis) using different light field super-resolution algorithms when considering a magnification factor of $\times 3$.}
\label{tbl:SR_psnr_analysis_x3}
\begin{center}
\begin{tabular}{|l|c|c|c|c|c|}
\hline
\bf{Light Field} & \bf{Bicubic} &   \bf{BM-PCARR} & \bf{LF-SRCNN} & \bf{GRAPH} & \bf{PB-VDSR} \\ 
\hline
Bikes          & 27.55 (0.87) & 28.78 (0.89) & 28.73 (0.88) & 29.22 (0.90) & 29.87 (0.90)\\
Bench in Paris & 22.43 (0.79) & 23.33 (0.83) & 23.07 (0.82) & 23.25 (0.83) & 23.48 (0.83)\\
Friends 1      & 31.17 (0.90) & 32.17 (0.92) & 32.17 (0.92) & 32.17 (0.92) & 33.16 (0.92)\\
Sphynx         & 27.65 (0.77) & 28.73 (0.81) & 28.45 (0.80) & 28.88 (0.81) & 28.76 (0.80)\\
Bee 2 		   & 31.02 (0.91) & 32.03 (0.91) & 32.24 (0.92) & 32.74 (0.93) & 32.62 (0.92)\\ 
Duck           & 23.35 (0.84) & 24.22 (0.86) & 24.19 (0.87) & 24.43 (0.88) & 24.50 (0.88)\\
Fruits         & 28.74 (0.85) & 30.21 (0.89) & 29.87 (0.88) & 30.91 (0.91) & 30.20 (0.89)\\
Rose           & 34.05 (0.90) & 35.30 (0.92) & 35.00 (0.91) & 36.19 (0.94) & 34.98 (0.91)\\
Mini           & 27.30 (0.77) & 28.23 (0.79) & 28.03 (0.79) & 28.31 (0.81) & 28.55 (0.80)\\
Chess          & 30.04 (0.92) & 31.02 (0.93) & 30.88 (0.93) & 31.69 (0.94) & 31.61 (0.94)\\
Bunny          & 32.91 (0.94) & 34.31 (0.94) & 34.14 (0.94) & 35.31 (0.96) & 35.64 (0.95)\\
Lego Bulldozer & 26.21 (0.86) & 27.05 (0.87) & 27.10 (0.87) & 28.27 (0.90) & 28.15 (0.89)\\
Lego Truck     & 30.26 (0.89) & 31.18 (0.91) & 30.99 (0.91) & 31.62 (0.92) & 31.39 (0.92)\\
Lego Knights   & 27.28 (0.86) & 28.15 (0.88) & 28.24 (0.87) & 28.62 (0.90) & 29.01 (0.90)\\
\hline
\hline
\bf{Overall} & \bf{ 28.57(0.86)} & \bf{29.62 (0.88)} & \bf{29.51 (0.88)} & \bf{30.12 (0.90)} & \bf{30.14 (0.89)}\\
\hline
\end{tabular}
\end{center}
\end{table*}

\begin{table*}[ht]
\caption{Quality analysis (PSNR with SSIM in parenthesis) using different light field super-resolution algorithms when considering a magnification factor of $\times 4$.}
\label{tbl:SR_psnr_analysis_x4}
\begin{center}
\begin{tabular}{|l|c|c|c|c|c|}
\hline
\bf{Light Field} & \bf{Bicubic} &  \bf{BM-PCARR} & \bf{LF-SRCNN} & \bf{GRAPH} & \bf{PB-VDSR} \\ 
\hline
Bikes          & 25.33 (0.80) & 26.42 (0.82) & 26.28 (0.82) & 26.62 (0.84) & 27.85 (0.82)\\
Bench in Paris & 21.00 (0.72) & 21.74 (0.75) & 21.50 (0.75) & 21.57 (0.75) & 21.73 (0.75)\\
Friends 1      & 29.15 (0.86) & 30.14 (0.88) & 30.10 (0.88) & 30.08 (0.88) & 30.89 (0.88)\\
Sphynx         & 25.88 (0.70) & 26.89 (0.74) & 26.62 (0.72) & 26.86 (0.74) & 26.79 (0.72)\\
Bee 2          & 28.72 (0.86) & 29.85 (0.87) & 29.85 (0.88) & 30.25 (0.89) & 30.27 (0.88)\\
Duck           & 21.62 (0.76) & 22.29 (0.79) & 22.25 (0.79) & 22.44 (0.81) & 22.50 (0.80)\\
Fruits         & 26.60 (0.78) & 27.82 (0.82) & 27.53 (0.80) & 28.28 (0.84) & 27.28 (0.80)\\
Rose           & 31.86 (0.84) & 33.05 (0.87) & 32.57 (0.85) & 33.42 (0.88) & 32.12 (0.84)\\
Mini           & 25.71 (0.70) & 26.40 (0.72) & 26.30 (0.71) & 26.45 (0.73) & 26.73 (0.72)\\
Chess          & 28.03 (0.87) & 28.90 (0.88) & 28.77 (0.88) & 29.31 (0.90) & 28.65 (0.88)\\
Bunny          & 30.47 (0.90) & 31.80 (0.91) & 31.57 (0.91) & 32.30 (0.92) & 32.33 (0.91)\\
Lego Bulldozer & 24.29 (0.79) & 25.00 (0.80) & 25.02 (0.81) & 25.85 (0.84) & 25.13 (0.81)\\
Lego Truck     & 28.55 (0.85) & 29.35 (0.87) & 29.15 (0.86) & 29.56 (0.87) & 29.09 (0.86)\\
Lego Knights   & 25.20 (0.79) & 26.13 (0.81) & 26.84 (0.81) & 26.67 (0.84) & 25.89 (0.81)\\
\hline
\hline
\bf{Overall} & \bf{26.60 (0.80)} & \bf{27.60 (0.82)} & \bf{27.45 (0.82)}& \bf{27.83 (0.84)} & \bf{27.66 (0.82)}\\
\hline
\end{tabular}
\end{center}
\end{table*}

\begin{figure*}[ht]
\centering
\setlength\tabcolsep{1.5pt}
\begin{tabular}{cccc}
\centering
\footnotesize{BM+PCARR} & \footnotesize{LF-SRCNN} &\footnotesize{GRAPH}& \footnotesize{PB-VDSR} 
\\ 
\includegraphics[width=4.3cm]{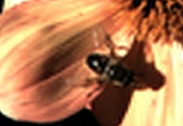} & 
\includegraphics[width=4.3cm]{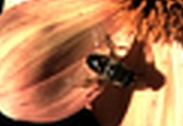}  & 
\includegraphics[width=4.3cm]{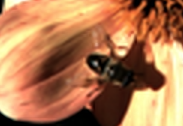}  & 
\includegraphics[width=4.3cm]{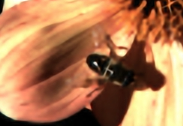}
\\
\includegraphics[width=4.3cm]{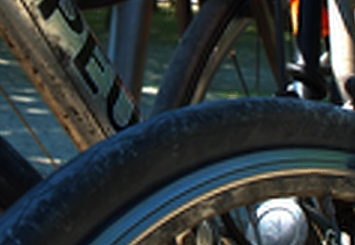} & 
\includegraphics[width=4.3cm]{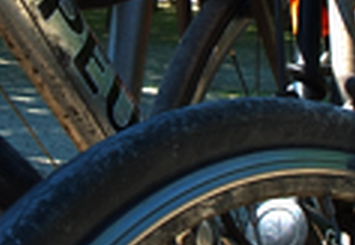}  & 
\includegraphics[width=4.3cm]{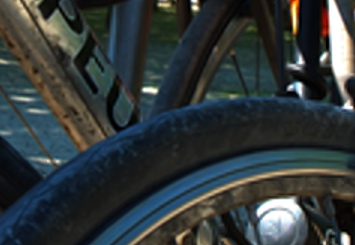}  & 
\includegraphics[width=4.3cm]{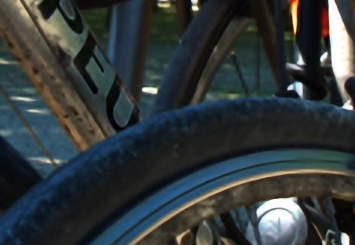}
\\
\includegraphics[width=4.3cm]{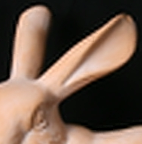} & 
\includegraphics[width=4.3cm]{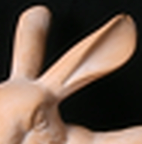}  & 
\includegraphics[width=4.3cm]{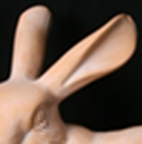}  & 
\includegraphics[width=4.3cm]{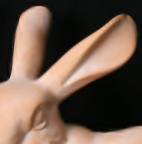}
\\
\includegraphics[width=4.3cm]{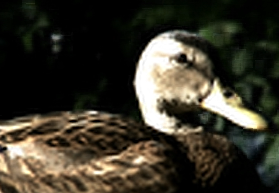} & 
\includegraphics[width=4.3cm]{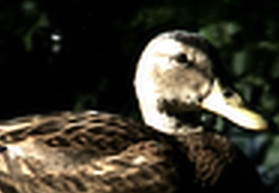}  & 
\includegraphics[width=4.3cm]{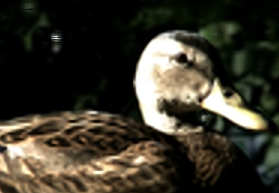}  & 
\includegraphics[width=4.3cm]{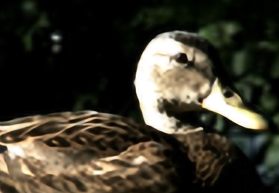}
\\
\includegraphics[width=4.3cm]{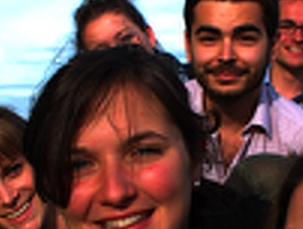} & 
\includegraphics[width=4.3cm]{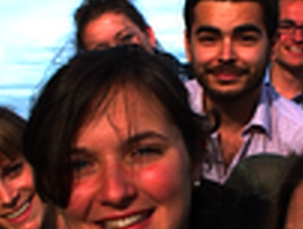}  & 
\includegraphics[width=4.3cm]{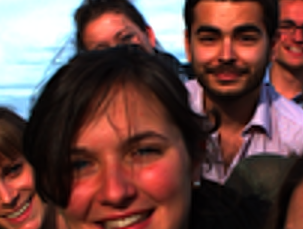}  & 
\includegraphics[width=4.3cm]{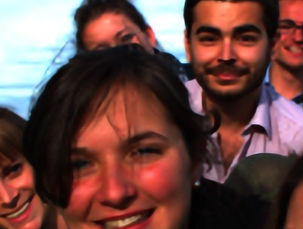}
\\
\includegraphics[width=4.3cm]{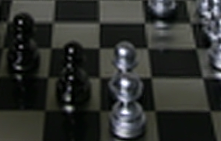} & 
\includegraphics[width=4.3cm]{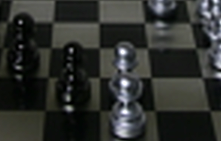}  & 
\includegraphics[width=4.3cm]{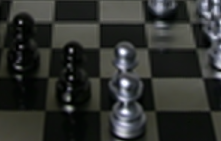}  & 
\includegraphics[width=4.3cm]{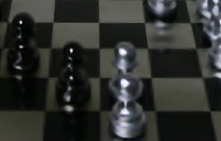}
\\
\includegraphics[width=4.3cm]{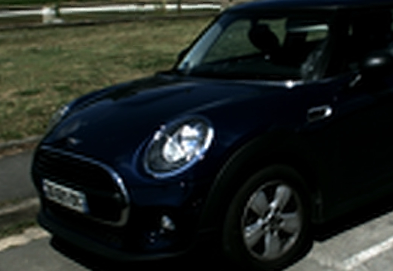} & 
\includegraphics[width=4.3cm]{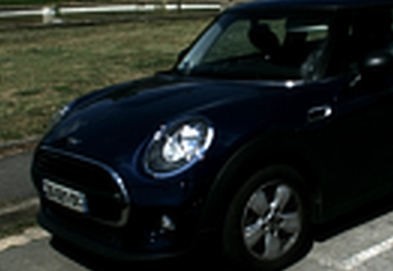}  & 
\includegraphics[width=4.3cm]{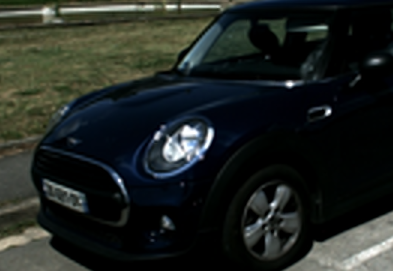}  & 
\includegraphics[width=4.3cm]{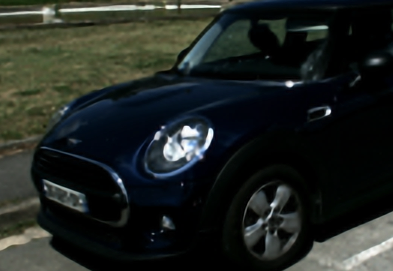}
\\
\end{tabular} 
\caption{Restored center view of light fields using different light field super-resolution algorithms. These are best viewed in color and by zooming on the views.}
\label{fig:SR_subjective_Eval}
\end{figure*}

The results in Table \ref{tbl:SR_psnr_analysis_x3} and Table \ref{tbl:SR_psnr_analysis_x4} compare these light field super-resolution methods in terms of both PSNR and SSIM for magnification factors of $\times3$ and $\times4$ respectively.
It can be seen that our proposed method outperforms both BM-PCARR and LF-SRCNN and it is competitive to the GRAPH light field super-resolution method when considering both PSNR and SSIM objective quality metrics.
Moreover, it can be seen in Figure \ref{fig:SR_subjective_Eval} that our method is able to restore central views that are much sharper and of higher quality (see bee in first row, text on the bicycle in second row, eyes of the duck in forth row and edges of the chess board in the sixth row of Figure \ref{fig:SR_subjective_Eval}) compared to the three leading light field super-resolution methods found in literature.
One can also notice that the other methods provide aliasing (see bicycle rim in second row and teeth of the second female on the left in the fifth row of Figure \ref{fig:SR_subjective_Eval}) and ghosting artifacts (see rabbit ears in third row in Figure \ref{fig:SR_subjective_Eval}).
It can also be seen that our proposed method yields sharper results on non-Lambertian surfaces as can be seen on the Mini light field (bottom row in Figure \ref{fig:SR_subjective_Eval}).

\begin{figure*}[ht]
\centering
\def\svgwidth{0.95\linewidth} 
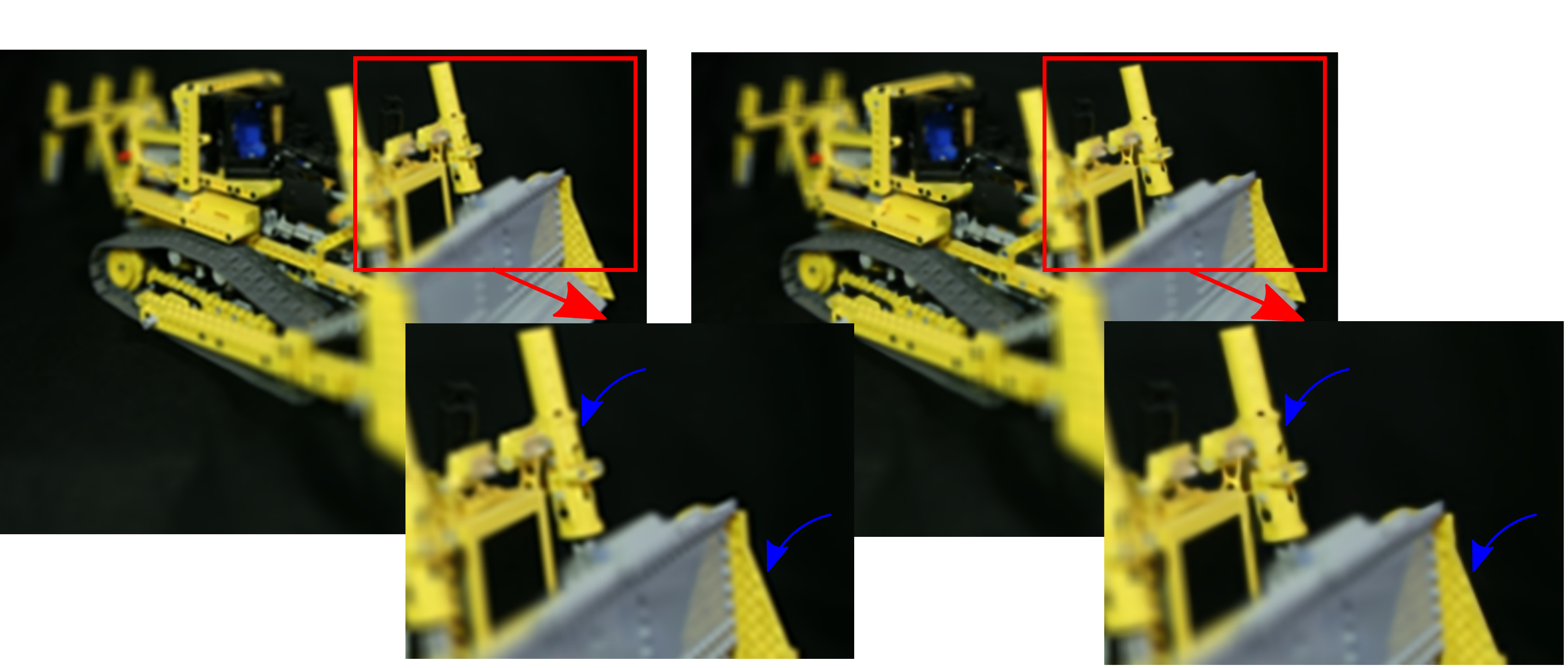
\\
\centerline{(a) Refocus of Lego Bulldozer at a slope of -0.6.}
\def\svgwidth{0.95\linewidth} 
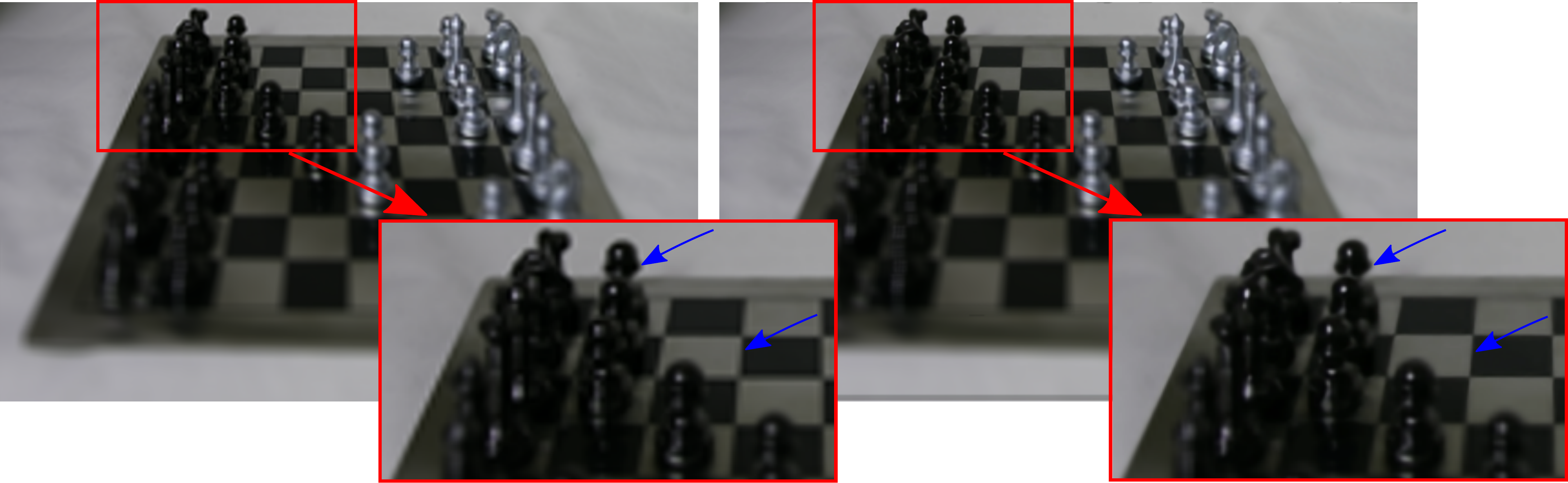
\\
\centerline{(b) Refocus of Chess at a slope of +0.1.}
\def\svgwidth{0.95\linewidth} 
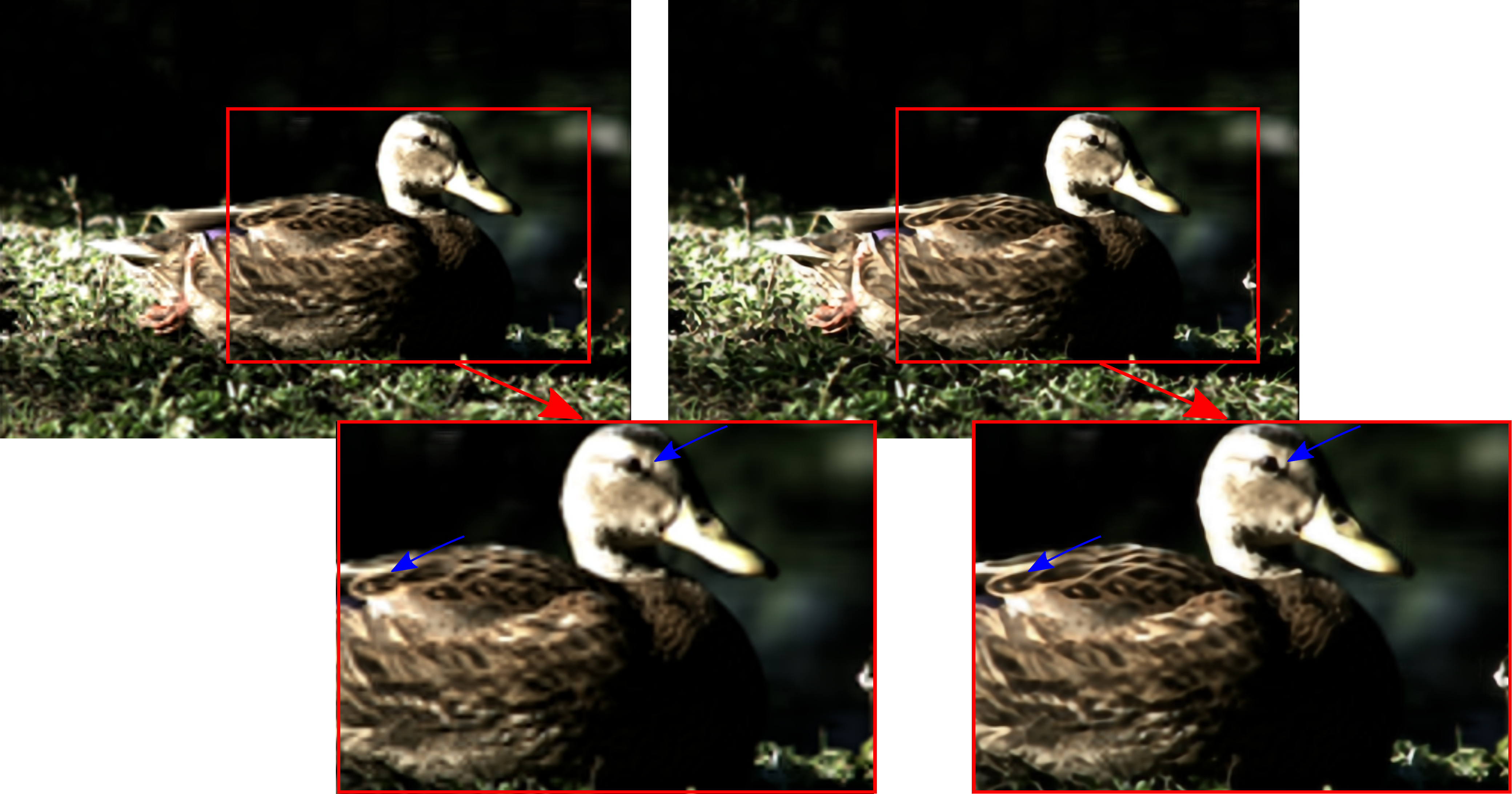
\\
\centerline{(c) Refocus of Duck at a slope of +0.0.}
\caption{Refocusing of different light field at different depths.}
\label{fig:refocus}
\end{figure*}

One important feature of a light field is that it enables to digitally refocus the image after production.
The quality of the refocused image depends on the quality of the light field and of its  coherence across all the views. 
The results in Figure \ref{fig:refocus} shows a number of refocused images obtained from light fields restored using the GRAPH \cite{Rossi2017} and PB-VDSR where the images were refocused using the Light Field Toolbox \cite{Dansereau2015}.
These results clearly show that the refocused images computed on light fields restored using PB-VDSR are sharper and of better quality.
Moreover, the supplementary multimedia files show a pseudovideo of different light fields reconstructed using PB-VDSR where it is evident that the proposed method manages to restore light fields of higher quality and better angularly coherence compared to those obtained using the GRAPH method, even when considering non-Lambertian surfaces as the Tarot Cards and Crystal Ball light field where the latter fails. 
Moreover, these result also show the restoration of real-world applications when super-resolving the plenoptic image from $625 \times 434$ to $1875 \times 1302$.

The complexity of PB-VDSR is mainly affected by the computation of the optical flows used to align the light field (SIFT Flow in our case), of the SVD decomposition used to decompose the aligned light field, of the single image super resolution method used to restore the \textit{principal basis} (VDSR in our case) and of the matrix multiplication that is used to propagate the restored information in the principle basis to all the other views.
The Sift Flow is used to align all the $n$ views to the center view and is reported in \cite{Liu2011} to have a time complexity of the order $O(nm\log (\sqrt{m}))$, where $m$ represents the number of pixels in each view.
The SVD decomposition and the matrix multiplication incur a time complexity of the order $O(n^2m)$ each.
Moreover, the feed-forward part of VDSR which is used during evaluation has a fixed depth and width and its complexity is mainly dependent on the resolution of the \textit{principal basis}.
This implies that the VDSR algorithm has a time complexity of the order $O(m)$.
This complexity analysis concludes that the proposed method has a time complexity that is mainly dependent on the resolution and number of views in the light field.
This contrasts with the GRAPH method presented in \cite{Rossi2017} whose time complexity is proportional to $\alpha^4$ where $\alpha$ is the magnification factor.
A quantitative assessment of the complexity of different light field super-resolution methods considered in this work is summarized in Table \ref{tbl:complexity}.
These methods were implemented using MATLAB with code provided by the authors and tested on an Intel Core(TM)i7 with a Windows 10 64-bit Operating System, 32-GByte RAM and a Titan GTX1080Ti GPU. The LF-SRCNN has the smallest time complexity. However, it registered the worst performance in terms of quality (see Tables \ref{tbl:SR_psnr_analysis_x3} and \ref{tbl:SR_psnr_analysis_x4}).
Our proposed method achieved the second lowest complexity which is clearly independent on the target magnification factor. 
On the other hand, the complexity of GRAPH is orders of magnitudes larger than our method and its complexity increases exponentially with increasing magnification factors.

\begin{table}[ht]
\caption{Processing time of different light field super-resolution algorithms at different magnification factors.}
\label{tbl:complexity}
\begin{center}
\begin{tabular}{|l|c|c|c|}
\hline
\bf{Algorithm} & $\times2$ &  $\times3$ & $\times4$  \\ 
\hline
BM-PCARR & 22 min.  & 23 min. & 23 min.\\
LF-SRCNN & 33 sec. & 33 sec. & 33 sec. \\
GRAPH    & 4 hrs. & 7 hrs.& 1 day \\
PB-VDSR & 9 min. & 9 min. & 9 min. \\
\hline
\end{tabular}
\end{center}
\end{table}

\section{Comments and Conclusion}
\label{sec:conclusion}

This paper has proposed a simple framework allowing to apply state-of-the-art SISR methods for light field super-resolution while preserving light field geometrical constraints. 
The problem is decomposed into two sub--problems where we first align each view to the center view using optical flows and we then decompose the aligned light field using SVD.
Experimental results show that the \textit{principal basis} captures the coherent information in the light field and is a natural image that can be restored using state-of-the-art SISR methods.
We also demonstrate that the information restored in the \textit{principal basis} can be propagated in a consistent manner to all the other views.
Experimental results show that the use of the VDSR SISR technique in the proposed framework manages to restore light fields that are sharper and coherent across the angular views, compared to existing light field super-resolution methods.
Moreover, results in the supplementary material show that the restored light field is able to restore reflections on non-Lambertian surfaces.
The proposed framework can be extended to other light field image processing applications such as inpainting and recolouring where one can edit the \textit{principal basis} using state-of-the-art 2D image processing methods, and then propagate the restored information to all the other views as it was done here.

\bibliographystyle{IEEEtran}
\bibliography{IEEEabrv,egbib}

\begin{IEEEbiography}[{\includegraphics[width=1in,height=1.25in,clip,keepaspectratio]{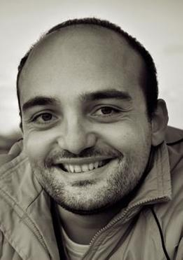}}]{Reuben A. Farrugia}
(S’04, M’09) received the first
degree in Electrical Engineering from the University of
Malta, Malta, in 2004, and the Ph.D. degree from the
University of Malta, Malta, in 2009.
In January 2008 he was appointed Assistant Lecturer with the same department
and is now a Senior Lecturer. 
He has been in technical and organizational committees of several national and international conferences. In particular, he served as General-Chair on the IEEE Int. Workshop on Biometrics and Forensics (IWBF) and as Technical Programme Co-Chair on the IEEE Visual Communications and Image Processing (VCIP) in 2014.  He has been contributing as a reviewer of several journals and conferences, including IEEE Transactions on Image Processing, IEEE Transactions on Circuits and Systems for Video and Technology and IEEE Transactions on Multimedia. On September 2013 he was appointed as National Contact Point of the European Association of Biometrics (EAB).
\end{IEEEbiography}

\begin{IEEEbiography}[{\includegraphics[width=1in,height=1.25in,clip,keepaspectratio]{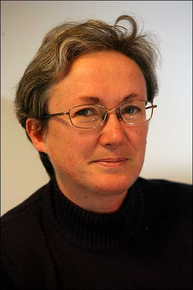}}] {Christine Guillemot} IEEE fellow, is “Director of Research” at INRIA, head of a research team
dealing with image and video modeling, processing, coding and communication. She holds a Ph.D. degree from ENST
(Ecole Nationale Superieure des Telecommunications) Paris, and an “Habilitation for Research Direction” from the
University of Rennes. From 1985 to Oct. 1997, she has been with FRANCE TELECOM, where she has been involved
in various projects in the area of image and video coding for TV, HDTV and multimedia. From Jan. 1990 to mid
1991, she has worked at Bellcore, NJ, USA, as a visiting scientist. She has (co)-authored 24 patents, 9 book chapters,
60 journal papers and 140 conference papers. She has served as associated editor (AE) for the IEEE Trans. on Image
processing (2000-2003), for IEEE Trans. on Circuits and Systems for Video Technology (2004-2006) and for IEEE Trans. On Signal Processing
(2007-2009). She is currently AE for the Eurasip journal on image communication, IEEE Trans. on Image Processing (2014-2016) and member of the editorial board for the IEEE Journal on
selected topics in signal processing (2013-2015). She is a member of the IEEE IVMSP technical committee.
\end{IEEEbiography}

\end{document}